\documentclass[10pt,twocolumn,letterpaper]{article}

\usepackage[pagenumbers]{cvpr} 

\usepackage{graphicx}
\usepackage{amsmath}
\usepackage{amssymb}
\usepackage{booktabs}

\usepackage{amsfonts}
\usepackage{xcolor}
\usepackage{multirow}
\usepackage[T1]{fontenc}

\usepackage[pagebackref,breaklinks,colorlinks]{hyperref}

\usepackage[capitalize]{cleveref}
\crefname{section}{Sec.}{Secs.}
\Crefname{section}{Section}{Sections}
\Crefname{table}{Table}{Tables}
\crefname{table}{Tab.}{Tabs.}

\begin{document}

\title{Unconstrained Face Mask \& Face Hand Datasets: Building a Computer Vision System to Help Prevent the Transmission of COVID-19
}

\author{Fevziye Irem Eyiokur \\
Karlsruhe Institute of Technology \\
{\tt\small irem.eyiokur@kit.edu}
\and
Hazım Kemal Ekenel \\
Istanbul Technical University \\
{\tt\small ekenel@itu.edu.tr}
\and
Alexander Waibel \\
Karlsruhe Institute of Technology \\
{\tt\small alexander.waibel@kit.edu}
}

\maketitle

\begin{abstract}
Health organizations advise social distancing, wearing face mask, and avoiding touching face to prevent the spread of coronavirus. Based on these protective measures, we developed a computer vision system to help prevent the transmission of COVID-19. Specifically, the developed system performs face mask detection, face-hand interaction detection, and measures social distance. To train and evaluate the developed system, we collected and annotated images that represent face mask usage and face-hand interaction in the real world. Besides assessing the performance of the developed system on our own datasets, we also tested it on existing datasets in the literature without performing any adaptation on them. In addition, we proposed a module to track social distance between people. Experimental results indicate that our datasets represent the real-world's diversity well. The proposed system achieved very high performance and generalization capacity for face mask usage detection, face-hand interaction detection, and measuring social distance in a real-world scenario on unseen data. The datasets will be available at \textcolor{magenta}{\href{https://github.com/iremeyiokur/COVID-19-Preventions-Control-System}{\textit{https://github.com/iremeyiokur/COVID-19-Preventions-Control-System}}}.

\end{abstract}

\section{Introduction}
\label{intro}

The COVID-19 pandemic has affected the whole world since the beginning of 2020. In order to decrease the transmission of the COVID-19 disease, many health institutions, particularly the World Health Organization (WHO), have recommended serious constraints and preventions \cite{WHO_website}. The fundamental precautions that individuals can carry out are to keep the distance from others (practicing social distance) \cite{WHO_website_distance}, wear a face mask properly (covering mouth and nose), pay attention to personal hygiene, especially hand hygiene, and avoid touching faces with hands without cleanliness \cite{WHO_website}.

Convolutional Neural Networks (CNNs), introduced in late 80s \cite{waibel1989phoneme,le1989handwritten}, have gained popularity during the last decade. Due to the success of deep learning in computer vision, novel research topics that emerged as a consequence of the COVID-19 pandemic are handled in this context by researchers. These studies focus on diagnosing COVID-19 \cite{chen2020deep,li2020using,farooq2020covid,narin2020automatic}, adjusting the already existing surveillance systems to COVID-19 conditions \cite{jiang2020retinamask,wang2020masked,anwar2020masked,damer2020effect,golwalkar2020age,chen2020efficient}, and building systems to control the preventions \cite{cabani2021maskedface,joshi2020deep,nagrath2021ssdmnv2,wang2020masked,batagelj2021correctly,chowdary2020face,wang2021wearmask,petrovic2020iot,loey2021fighting,loey2021hybrid,sathyamoorthy2020covid,yang2020vision,rezaei2020deepsocial,ahmed2021deep}. While some of the studies employ CT scans \cite{chen2020deep,li2020using} to diagnose COVID-19, the others benefit from chest X-ray images \cite{farooq2020covid,narin2020automatic}. Face detection and recognition systems' performance deteriorates when subjects wear face masks. Thus, novel face recognition and detection studies \cite{jiang2020retinamask,anwar2020masked,damer2020effect} try to improve the performance under this condition. Besides, the age prediction \cite{golwalkar2020age} is investigated when face mask is used. Moreover, in order to track the execution of preventions against the spread of COVID-19, several works investigate the detection of wearing a mask suitably \cite{cabani2021maskedface,joshi2020deep,nagrath2021ssdmnv2,wang2020masked,batagelj2021correctly,chowdary2020face,wang2021wearmask,petrovic2020iot,loey2021fighting,loey2021hybrid} and keeping the distance from other people \cite{petrovic2020iot,sathyamoorthy2020covid,yang2020vision,rezaei2020deepsocial,ahmed2021deep}.

In \cite{wang2020masked}, a novel masked face recognition dataset is published to improve the face recognition performance in the case of occlusion due to face masks. This dataset contains three subsets, which are Masked Face Detection Dataset (MFDD), Real-world Masked Face Recognition Dataset (RMFRD), and Simulated Masked Face Recognition Dataset (SMFRD). In \cite{cabani2021maskedface}, an artificial masked face dataset, named as MaskedFace-Net is presented. It contains 137016 images that are generated from the FFHQ dataset \cite{karras2019style} using a mask-to-face deformable model. Joshi et. al \cite{joshi2020deep} proposed a framework to detect whether people are wearing a mask or not in public areas. They utilized MTCNN \cite{mtcnn_paper} and MobileNetV2 \cite{mobilenetv2model} to detect faces and classify them on their own video dataset. In \cite{jiang2020retinamask}, a one-stage detector based on RetinaFace \cite{Deng_2020_CVPR} is proposed to detect faces and classify them whether they contain masks. In \cite{nagrath2021ssdmnv2}, the authors proposed a real-time face mask detector framework named SSDMNV2, which is composed of SSD \cite{liu2016ssd} as a face detector and MobileNetV2 \cite{mobilenetv2model} as a mask classifier. A recent study \cite{beyan2020analysis} investigated the face-hand touching behavior. In this study, the authors presented face-hand touching interaction annotations on 64 video recordings and they evaluated the annotated 2M non-touching and 74K touching frames with rule-based, hand-crafted feature-based, and CNN learned feature-based models. As a result of evaluations, CNN based model obtained the best results with 83.76\% F1-score.

In this work, we present a computer vision system that controls preventions advised by the health institutions. These preventions are to detect whether people wear a face mask, keep away from touching their faces, and to monitor the social distance. To train and evaluate the developed system, we collected two novel face datasets, namely Interactive Systems Labs Unconstrained Face Mask Dataset (ISL-UFMD) and Interactive Systems Labs Unconstrained Face Hand Interaction Dataset (ISL-UFHD). These datasets are collected from the web to provide a significant amount of variation in terms of pose, illumination, resolution, and ethnicity. The system consists of three submodules, corresponding to face mask detection, face-hand interaction detection, and social distance measurement tasks, respectively. We trained two separate CNN models to classify face images for the face mask and face-hand interaction detection tasks. While the first model classifies the face image as wearing a mask properly, wearing a mask improperly, or not wearing a mask, the second model classifies face images as touching the face or not. The trained models were evaluated both on the collected dataset and on the existing face mask datasets in the literature without training or fine-tuning on them.  We also proposed an approach to measure the social distance. Our contributions can be summarized as follows: (1) We develop a computer vision system to help people to follow the recommended protective measures --wearing a face mask properly, not touching faces, having social distance-- to avoid spread of COVID-19. (2) We present two novel datasets, ISL-UFMD and ISL-UFHD, for face mask and face-hand interaction detection tasks. ISL-UFMD is one of the largest face mask datasets that includes images with a significant amount of variations. The ISL-UFHD is the first dataset that contains face-hand interaction images from unconstrained real-world scenes. (3) We extensively investigate several CNN models on our datasets. We also tested them on publicly available masked face datasets without performing adaptation, e.g. fine-tuning, on them to demonstrate the generalization capacity of our models. We achieved very high classification accuracies which indicates the collected datasets' capability to represent real-world cases. Moreover, to evaluate the overall system, we utilized six different short real-world video recordings.

\begin{table*}
\begin{center}
\setlength{\tabcolsep}{3pt}
\caption{Comparison of the face mask datasets. (*) Although it is stated that RMFD dataset \cite{yang2020vision} contains 5000 face images with mask, there are only 2203 face images with mask in the publicly available version. }
\label{dataset_information}       
\begin{tabular}{lcccccccc}
\hline\noalign{\smallskip}
Dataset name & No mask & Mask & Improper Mask & Data Type & Ethnicities & Head Pose\\
\noalign{\smallskip}\hline\noalign{\smallskip}
ISL-UFMD & 10698 & 10618 & 500 & Real &  Various & Various \\
\noalign{\smallskip}\hline\noalign{\smallskip}
RMFD \cite{yang2020vision}* & 90468 & 2203 & - & Real & Asian & Frontal to Profile \\
RWMFD \cite{yang2020vision} & 858 & 4075 & 238 & Real & Mostly Asian & Frontal to Profile \\
Face mask \cite{kaggle853}  & 718 & 3239 & 123 & Real &   Mostly Asian & Various \\
MaskedFace-Net \cite{cabani2021maskedface} & - & 67049 & 66734 & Artificial &  Various & Mostly Frontal \\
\noalign{\smallskip}\hline
\end{tabular}
\end{center}
\end{table*}

\section{The ISL-UFMD \& ISL-UFHD Datasets}
\label{dataset}
To train our system, we collected both face masked images and face-hand interaction images. Recently published datasets on the tracking of COVID-19 preventions, which are presented in Table \ref{dataset_information}, mainly focused on collecting face mask images to develop a system that examines whether there is a mask on the face or not. Most of them contain a limited amount of images or include synthetic images generated with putting a mask on the face using landmark points around the mouth and nose. Besides, the variety of subjects' ethnicity, image conditions, such as environment, resolution, and particularly different head pose variations are limited. For instance, in these datasets except MaskedFace-Net \cite{cabani2021maskedface}, Asian people are in the majority. Although MaskedFace-Net includes variation in terms of ethnicity, it consists entirely of images with artificial face masks. Besides, all face mask datasets have limited head poses mostly from frontal view to profile view in yaw axis. Thus, these limitations led us to collect a dataset to overcome all these drawbacks. In addition to face mask, there is only one dataset \cite{beyan2020analysis} that is recently annotated to investigate face-hand interaction in the literature. However, these face-hand interaction annotations are also limited based on the number of subjects. Moreover, the dataset is collected in an indoor environment under controlled conditions. In this study on the other hand, we present a face-hand interaction dataset that is collected from unconstrained real world scenes.

\subsection{Data Collection} 
We collected a large amount of face images from several different resources, such as publicly available face datasets, FFHQ \cite{karras2019style}, CelebA \cite{celebahq}, LFW \cite{LFW}, YouTube videos, and web. These various sources enable us to collect a significant variety of face images in terms of ethnicity, age, and gender. In addition to the subject diversity, we obtained images from indoor and outdoor environments, under different light conditions and resolutions. We also considered ensuring large head pose variations. Moreover, another important key point is to leverage the performance of our COVID-19 prevention system for the combined scenario, e.g., determining mask usage in the case of touching faces or detecting face-hand interaction in the case of wearing a mask. Besides, our images include different sorts of occlusion that make the dataset more challenging. In the end, ISL-UFMD contains 21316 face images for the face-mask detection scenario, 10618 face images with masks and 10698 images without a mask. Additionally, we gathered 500 images for improper mask usage. This class has a relatively small number of images compared to no mask and mask classes due to lack of face images with improper mask usage.

\begin{figure}
\begin{center}
  \includegraphics[width=0.46\textwidth]{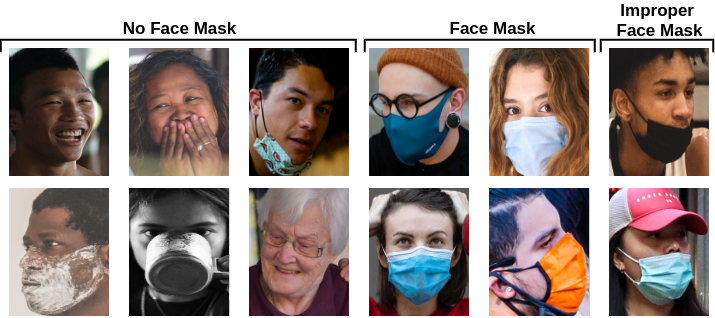}
\caption{Example images from ISL Unconstrained Face Mask Dataset (ISL-UFMD). This figure shows samples from three different classes; no mask, face mask, improper face mask.}
\label{fig_sample_mask}       
\end{center}
\end{figure}

\begin{figure}
\begin{center}
  \includegraphics[width=0.46\textwidth]{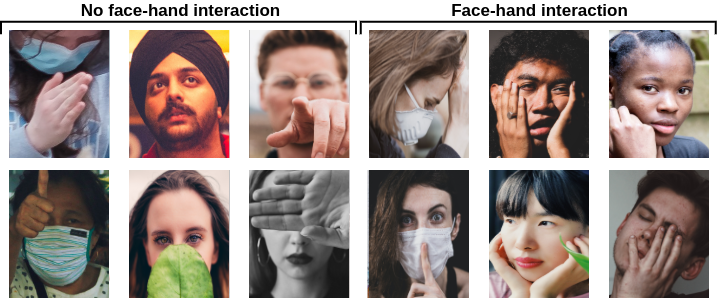}
\caption{Example images from ISL Unconstrained Face Hand Interaction Dataset (ISL-UFHD). This dataset contains images that represent face-hand interaction and no interaction.}
\label{fig_sample_hand}       
\end{center}
\end{figure}

The ISL-UFHD is composed of face images that represent the interaction between the face and hand of the subjects. We collected 22289 negative samples (no face-hand interaction) and 10004 positive samples (face-hand interaction). Please note that, even if the hand is around the face without touching it, we annotated it as a no interaction. Therefore, the model should be able to distinguish whether the hand in the image is touching the face (or very close to the face) or not.

\subsection{Data Annotation}

\begin{figure*}
\begin{center}
  \includegraphics[width=0.99\textwidth]{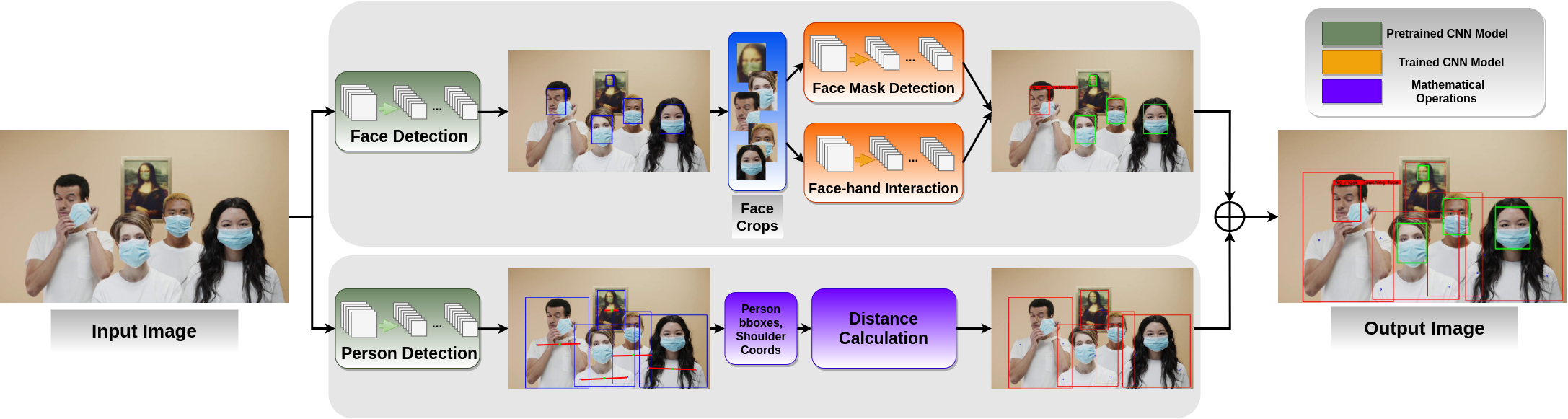}
\caption{Proposed system for controling COVID-19 preventions.}
\label{fig_system}       
\end{center}
\end{figure*}

For labelling the collected datasets, we designed a web-based image annotation tool. We utilized crowd-sourcing to annotate each image and after examining these annotations, we decided each image's final label. Since we formulated our tasks as classification problems, we annotated our images in that manner. While we have three classes --mask, no mask, improper mask-- for the mask detection task, we have two classes for the face-hand interaction detection task. The images that include the face without a fully covered nose and mouth by the mask are annotated with the improper mask label. If a person has a mask under the chin, we annotated the image with the no mask label. In the face-hand annotation, we considered the direct contact or too close to contact as the existence of face-hand interaction. Many examples of annotated face images for face mask and face-hand interaction detection tasks are shown in Fig.~\ref{fig_sample_mask} and Fig. \ref{fig_sample_hand}. It can be clearly seen from the figures that our proposed datasets contain large amount of variations especially for gender, ethnicity, and head pose. Also, the examples have diversity in terms of position of hand upon face and usage of face mask.

\section{Methodology}
\label{methodology}

The proposed COVID-19 prevention control system is illustrated in Fig. \ref{fig_system}. The proposed system consists of three submodules. Each module utilizes deep CNN models to obtain predictions. The system performs person detection  and calculates distances between detected subjects on input image/video frame. Meanwhile, the same input is also used to detect and crop faces of subjects to perform face mask and face-hand interaction detections. While the face mask model decides whether a person wears a mask (properly) or not, the face-hand interaction model identifies whether a hand touches the subject's face. We decided to perform person detection and face detection separately on the input image/video frame to eliminate the effect of missing modality. For instance, although a person's body is occluded and social distancing cannot be measured with this person, system can still detect the face of the corresponding subject to perform face mask and face-hand interaction tasks. Similarly, if the subject's face is occluded or not turned to the camera, system can be able to capture the person's body to perform the social distance task.

\subsection{Face mask and face-hand interaction detection}

For face mask and face-hand interaction tasks, firstly, we performed face detection using the pretrained ResNet50 \cite{resnetmodel} backboned RetinaFace model \cite{Deng_2020_CVPR} that was trained on the large-scale Wider-Face dataset \cite{yang2016wider}. We used RetinaFace detector since it is robust against tiny faces, challenging head poses, and faces with a mask. After detection, we cropped detected faces with a 20\% margin for each side, since the face detector's outputs are quite tight. To perform face mask and face-hand interaction detections, we employed several different CNN architectures, namely, ResNet50 \cite{resnetmodel}, Inception-v3 \cite{inceptionv3model}, MobileNetV2 \cite{mobilenetv2model}, and EfficientNet \cite{efficientnetmodel}. We decided to use EfficientNet, since it is the state-of-the-art model. We also included MobileNetV2, since it is a light-weight deep CNN model. Finally, we chose ResNet and Inception-v3 models based on their high performances in the literature. In the training, we benefited from transfer learning and initialized our networks with the weights of the pretrained models that were trained on ImageNet dataset \cite{deng2009imagenet}. We employed softmax loss at the end of each network. In EfficientNet and MobileNetV2, we utilized dropout with a 0.2 probability rate to avoid overfitting. We addressed the mask classification task as a multi-class classification \textit{--improper mask, proper mask, no mask--} problem. We handled the face-hand interaction detection as a two class classification \textit{--interaction, no interaction--} task. We aim to identify whether the hand touches the face using 2D images without using predefined or estimated depth information. At first, the input data passes through the face detector to detect bounding box coordinates of the faces. Then, these are used to obtain face crops with suitable margins. Afterward, the face mask and face-hand interaction models are used to predict on acquired face crops.

\begin{table*}
\begin{center}
\caption{Face mask detection results on proposed ISL-UFMD dataset for three classes case.}
\label{mask_classification_3}   
\begin{tabular}{lccccccc}
\hline\noalign{\smallskip}
\multicolumn{1}{l}{\multirow{2}{*}{Model}} & \multicolumn{1}{c}{\multirow{2}{*}{Accuracy}} & \multicolumn{3}{c}{Precision} & \multicolumn{3}{c}{Recall} \\
\multicolumn{1}{c}{}                       & \multicolumn{1}{c}{}                          & No Mask        & Mask        & Improper Mask        & No Mask       & Mask       & Improper Mask      \\
\noalign{\smallskip}\hline\noalign{\smallskip}
Inception-v3                               & \textbf{98.20\%}                                       &          0.985         &    0.986     &    0.833     &    0.988     &    0.984  & 0.800      \\
ResNet50                                  & 95.63\%                                       &         0.965 &    0.954      &   0.636      &  0.973       &  0.973       &   0.389     \\
MobileNetV2                                & 97.91\%                                       &         0.988 &   0.975       &  0.842       &   0.983      &  \textbf{0.992}       &   0.640     \\
EfficientNet-b0                            & 97.82\%                                       &         0.973 &  0.984        &  \textbf{0.929}       &  \textbf{0.992}       &  0.986       &   0.520     \\
EfficientNet-b1                            & 97.91\%                                       &         0.979 &   0.986       &   0.800      &    0.990     &    0.984     &  0.711      \\
EfficientNet-b2                            &   97.91\%                                            &   \textbf{0.990}       &  0.977        & 0.792        &     0.977    &     \textbf{0.992}    &  0.760      \\
EfficientNet-b3                            &    98.19\%                                           &    0.988      &    \textbf{0.990}      &  0.733       &  0.986       &  0.982       & \textbf{0.880}  \\
\noalign{\smallskip}\hline
\end{tabular}
\end{center}
\end{table*}

\subsection{Social distance controlling}
 
Keeping the social distance from others is another crucial measurement to avoid spreading of COVID-19.
For this, firstly, we detect each person on the image using a pretrained person detection model, DeepHRNet \cite{deepHRNet}. Thus, we obtain bounding boxes around the people and estimated pose information of each person.
Principally, we focus on the shoulders' coordinates to measure the approximate body width of a person on the projected image. In many studies, measurements are calculated based on the bounding box around the person. However, when the angle of the body joints and pose of the person are considered, changes on the bounding boxes may reduce the precision of the measurements. To prevent this, we propose to use shoulders' coordinates to measure the width of the body and identify the middle point of shoulders line as center of the body. 
After performing detection and pose estimation, we generated pairs based on the combination of the detected persons, e.g., $ P(p_i, p_j) $. Then, we calculated the Euclidean distance between the centers of shoulder points of each pair of persons.  
In order to decide whether these persons keep social distance between each other, we adaptively calculate a threshold for each pair individually based on the average of their body width. Since the represented measurement of the real world, expressed by pixels in the image domain, constantly changes as depth increases, we represent the mapping between real-world and pixel domain measurements by calculating the average of the body widths of two people to express this effect. Since the average distance between shoulder points of an adult is around 40-50 cm in the real-world and the required social distance between two persons is 1.5-2.0 meters, we empirically decide to select $\lambda$ coefficient as 3 when calculating the threshold for social distance in the pixel domain as in Eq. \ref{equation_threshold}.

\begin{equation}
\label{equation_threshold}
T_{p_i, p_j} = \frac{\lambda \times (||p_{i_{s_1}} - p_{i_{s_2}}||_2 + ||p_{j_{s_1}} - p_{j_{s_2}}||_2)}{2}
\end{equation}

\begin{table*}
\begin{center}
\caption{Results for cross-dataset experiments. All models are trained and tested on corresponding dataset. Please note that all experiments are conducted on the 3-class classification setup to perform fair comparison.}
\label{table_comparison}       
\begin{tabular}{lccccc}
\hline\noalign{\smallskip}
\multicolumn{1}{l}{\multirow{2}{*}{Architecture}} & \multicolumn{1}{c}{\smallskip\multirow{2}{*}{Training Set}} & \multicolumn{1}{c}{\multirow{2}{*}{Test Set}} & \multicolumn{2}{c}{\# Images } & \multicolumn{1}{c}{\multirow{2}{*}{Accuracy}} \\
& & & \multicolumn{1}{c}{} Train & \multicolumn{1}{c}{} Test & \\
\noalign{\smallskip}\hline\noalign{\smallskip}
MobileNetV2 & ISL-UFMD & RMFD \cite{wang2020masked} & 20764 & 92671 & 91.4\% \\
MobileNetV2 & ISL-UFMD & RWMFD \cite{wang2020masked} & 20764 & 5171 & 94.7\% \\
MobileNetV2 & ISL-UFMD & MaskedFace-Net \cite{cabani2021maskedface} & 20764 & 133782 & 88.11\% \\
MobileNetV2 & ISL-UFMD & Face-mask \cite{kaggle853} & 20764 & 4080 & \textbf{95.71\%} \\
Inception-v3 & ISL-UFMD & RMFD \cite{wang2020masked} & 20764 & 92671 & \textbf{95.91\%} \\
Inception-v3 & ISL-UFMD & RWMFD \cite{wang2020masked} & 20764 & 5171 & \textbf{95.9\%} \\
Inception-v3 & ISL-UFMD & MaskedFace-Net \cite{cabani2021maskedface} & 20764 & 133782 & \textbf{91.42\%} \\
Inception-v3 & ISL-UFMD & Face-mask \cite{kaggle853} & 20764 & 4080 & 94.7\% \\
\noalign{\smallskip}\hline\noalign{\smallskip}
MobileNetV2 & RMFD + RWMFD & ISL-UFMD & 97842 & 21816 & 86.59\% \\
MobileNetV2 & RMFD + RWMFD & Face-mask\cite{kaggle853} & 97842 & 4080 & 91.07\% \\
MobileNetV2 & MaskedFace-Net + FFHQ & ISL-UFMD & 211936 & 21816 & 51.49\% \\ 
MobileNetV2 & MaskedFace-Net + FFHQ & Face-mask\cite{kaggle853} & 211936 & 4080 & 20.4\% \\
Inception-v3 & RMFD + RWMFD & ISL-UFMD & 97842 & 21816 & 88.92\% \\
Inception-v3 & RMFD + RWMFD & Face-mask\cite{kaggle853} & 97842 & 4080 & 88.4\% \\
Inception-v3 & MaskedFace-Net + FFHQ & ISL-UFMD & 211936 & 21816 & 51.39\% \\ 
Inception-v3 & MaskedFace-Net + FFHQ & Face-mask\cite{kaggle853} & 211936 & 4080 & 19.2\% \\
\noalign{\smallskip}\hline
\end{tabular}
\end{center}
\end{table*}

Finally, if the Euclidean distance between two persons is lower than the calculated threshold, we decide that these people do not keep sufficient social distance.

\section{Experimental Results}
\label{experimental_results}

We used publicly available datasets to evaluate the generalization capacity of our system and also compared our mask detection models with the previous works. We used RMFD and RWMFD \cite{wang2020masked}\footnote{https://github.com/X-zhangyang/Real-World-Masked-Face-Dataset}. In RMFD, the publicly available version includes around 2203 masked face images, although the paper indicates that there are 5000 face mask images. For RWMFD, we executed RetinaFace and we obtained 5171 face images from 4343 images. MaskedFace-Net dataset \cite{cabani2021maskedface} contains ~130000 face images with artificial masks. While the half of the dataset (CMFD) has correctly worn face masks, the remaining half (IMFD) has incorrectly worn face masks. Face-mask dataset (Kaggle) \cite{kaggle853} contains 853 images and we used provided annotations to crop face images and obtain labels. In the end, we acquired 4080 face images. 

We used 90\% of the data for training, the remaining data is reserved equally for validation and testing. We followed the same strategy for face-hand interaction dataset. Additionally, before creating train-validation-test splits, we put aside around 5000 images from no face-hand interaction class to obtain a balanced dataset to execute face-hand interaction detection.

\subsection{Face mask detection} 

In Table \ref{mask_classification_3}, we presented various evaluation results using classification accuracy, precision, and recall. In the table, while the first column indicates the employed CNN models, the following columns represent evaluation results for face mask detection with these models. According to the experimental results in Table \ref{mask_classification_3}, although all employed models achieved significantly high performance, the best one is Inception-v3 model with 98.20\% classification accuracy. In addition to the classification accuracy, we also present precision and recall measurements for each class separately to demonstrate the performance of the models individually. In Table~\ref{mask_classification_3}, although the precision and recall values are very accurate for no mask and mask classes, these results for improper mask class are slightly lower than these two classes. Even though improper face mask can be confusing in terms of discrimination from mask class (proper), the more probable reason behind this outcome is the lack of images for improper mask usage. 

In Fig. \ref{cam}a, we present Class Activation Maps (CAM) \cite{selvaraju2017grad} for the face mask detection task to investigate activation of the model. It is clearly seen that the model focuses on the middle part of the faces, particularly on the nose and mouth. In the second image, the model identified improper mask usage since the nose of the subject is not covered by the face mask even though the mouth is covered. In Fig. \ref{cam}c, we present some misclassified images for the face mask detection task. Although the model classified the images incorrectly, the prediction probabilities of the model are not as high as in correct predictions. This outcome indicates that the model did not confidently misclassify images. Still, the difficulty in the head pose, illumination conditions, occlusion caused misclassification in some cases.

\paragraph{Cross-dataset experiments} In Table \ref{table_comparison}, we presented cross-dataset experiments to investigate the effect of the datasets on the generalization capacity of the proposed model. First, we evaluated our MobileNetV2 and Inception-v3 models on four different public face mask datasets. Additionally, we finetuned the MobileNetV2 and Inception-v3 models with two different training setups to compare our approach. The first setup contains 97842 images from the combination of RMFD and RWMFD datasets \cite{wang2020masked}. We used them together since RMFD dataset has no improper mask class. The second setup includes 211936 images from the MaskedFace-Net dataset \cite{cabani2021maskedface} with FFHQ dataset \cite{karras2019style}. We used FFHQ dataset as a no mask data due to the absence of no mask class on MaskedFace-Net dataset. While we selected RMFD, RWMFD, MaskedFace-Net, and Face-mask (Kaggle) \cite{kaggle853} datasets as target for our model, we used the proposed ISL-UFMD dataset and Face-mask (Kaggle) dataset as target datasets for other models. Almost all the models that are trained on the ISL-UFMD achieved more than 90\% accuracy. These results indicate that our ISL-UFMD dataset is significantly representative to provide well generalized models for face mask detection task. We employed two different architectures to endorse this outcome. Otherwise, the combination of RMFD and RWMFD provide accurate results, although they are not as high as our results. On the contrary, the models that are trained on the MaskedFace-Net dataset show the worst performance. The possible reason of this outcome is that the artificial dataset is not as useful as the real data for training. 

\begin{table}
\begin{center}
\caption{Face-hand interaction detection results on proposed ISL-UFHD dataset.}
\label{hand_classification}  
\begin{tabular}{lccc}
\hline\noalign{\smallskip}
\multicolumn{1}{l}{Model} & Accuracy & \multicolumn{1}{l}{Precision} & \multicolumn{1}{l}{Recall} \\
\noalign{\smallskip}\hline\noalign{\smallskip}
Inception-v3              & 93.20\%  &         0.932                      &         0.932                   \\
ResNet50                 & 91.76\%  &        0.918                       &         0.918                   \\
MobileNetV2               & 92.37\%  &          0.924                     &         0.924                   \\
EfficientNet-b0           & 92.37\%  &          0.926                     &          0.924                 \\
EfficientNet-b1           & 92.90\%  &         0.929                      &           0.929                \\
EfficientNet-b2           & \textbf{93.35\%}  &         \textbf{0.933}                      &           \textbf{0.934}                \\
EfficientNet-b3           & 92.44\%  &          0.925                     &             0.924               \\
\noalign{\smallskip}\hline
\end{tabular}
\end{center}
\end{table}

\subsection{Face-hand interaction detection} 

In Table \ref{hand_classification}, we present the face-hand interaction detection results. As in the face mask detection task, all of the employed models have achieved very high performance to discriminate whether there is an interaction between face and hand. The best classification accuracy is obtained as 93.35\% using EfficientNet-b2 model. The best recall and precision results are achieved by EfficientNet-b2 model as well. However, almost all results in the table is considerably similar to each other. Precision and recall metrics are balanced and compatible with the accuracies.

\begin{figure}
\small
\begin{center}
\begin{tabular}{cc}
\includegraphics[width=0.42\textwidth]{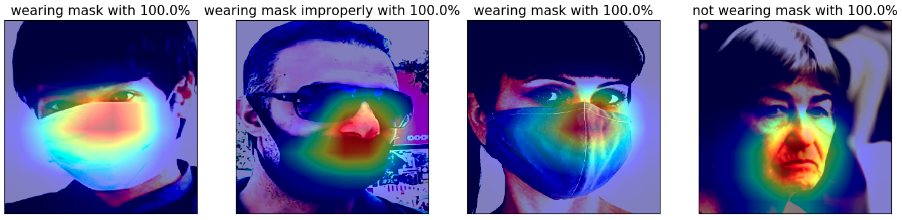}
& \\
(a)& \\
\includegraphics[width=0.42\textwidth]{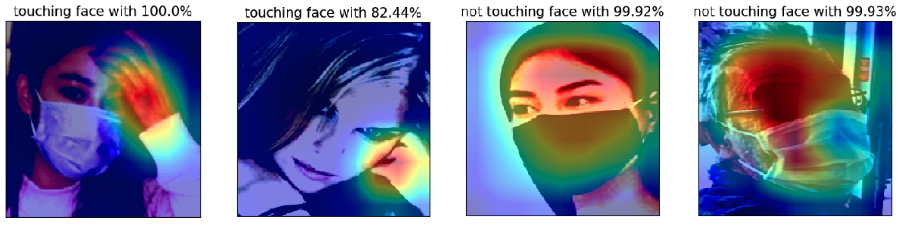}
& \\
(b)& \\
\includegraphics[width=0.42\textwidth]{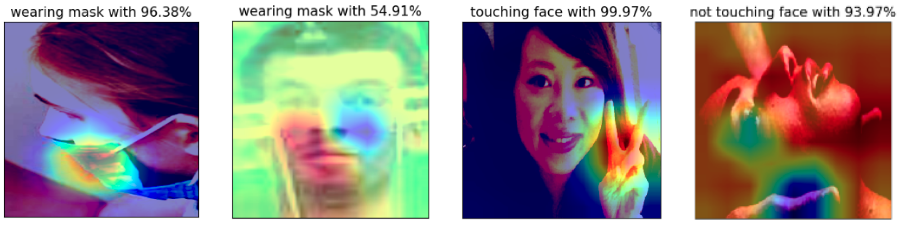}
\\
(c) \hspace{3.6cm} (d) & \\
\end{tabular}
\end{center}
\caption{Class activation map (CAM): (a) face mask detection task, (b) face-hand interaction detection task, (c) misclassified samples of face mask detection task, (d) misclassified samples of face-hand interaction detection task}
\label{cam}
\end{figure}

In Fig. \ref{cam}b, we provide CAM \cite{selvaraju2017grad} for the face-hand interaction detection. It is clearly seen that the model focuses on the hand and around the hand to decide whether there is an interaction between the hand and the face. In Fig. \ref{cam}d, we present some misclassified images for the face-hand interaction detection. In the first image, although the model can detect the hand and the face, it cannot identify the depth between the face and the hand due to the position of the hand. In the second image, it is seen that there is an interaction between the face and hands of someone else. For this example, the angles of the head and hands are challenging.

\subsection{Social distance controlling}

We utilized six different videos that we collected from the web to evaluate proposed social distancing control module. These videos have different number of frames and they were recorded in various environments with different camera angles. During the calculation of the accuracy of the social distance measurement algorithm, we utilized the annotations that we decided based on subject pairs and existing distance between each other. Person detector could not detect some of the subjects in the scene, if they are not visible in the camera due to the occlusion by other people or objects. For that reason, we ignored the missing detections when we annotated the videos' frames and calculated the accuracies. According to the results in Table \ref{table_overall}, we achieved very high accuracies on average. However, the fundamental problem, especially occurred in the last video, is caused by the lack of depth information.We project real-world distances to the image pixels with a rule-based approach without using reference points. Therefore, depth perception can be problematic for specific angles.

\begin{table}
\begin{center}
\setlength{\tabcolsep}{1.4pt}
\caption{Evaluation of the overall system on the test videos.}
\label{table_overall}       
\begin{tabular}{lccccc}
\hline\noalign{\smallskip}
Video & \# frames & \# sub. & Mask acc. & Face-hand acc. & Dist. acc. \\
\noalign{\smallskip}\hline\noalign{\smallskip}
V1 & 179  & 2 & 100\% & 99.16\% & 98.32\% \\
V2 & 307 & 2 & 99.51\% & 96.25\% & 100\% \\
V3 & 303 & 3 & 96.91\% & 89.43\% & 96.69\% \\
V4 & 192 & 3 & 100\% & 86.97\% & 97.22\% \\
V5 & 207 & 5 & 99.03\% & 95.45\% & 100\% \\
V6 & 105 & 7 & 87.07\% & 99.86\% & 74.55\% \\
\noalign{\smallskip}\hline\noalign{\smallskip}
Total & 1293 & 22 & 97.95\% & 93.84\% & 96.51\% \\
\hline
\end{tabular}
\end{center}
\end{table}

\subsection{Overall system performance}

We evaluated the overall system performance on the same six videos and presented the results in Table \ref{table_overall}. When we examined the face-hand interaction and face mask detection performance of our system, the results on videos that contains various people and cases indicate that system can reach very high performance similar to the ones that are obtained by the models on individual test sets.

\section{Conclusion}
\label{Conclusion}

In this paper, we proposed a system to track essential COVID-19 preventions. We collected two unconstrained datasets, ISL-UFMD and ISL-UFHD, with high diversity. While we employed several different CNN-based models to perform face mask and face-hand interaction detection tasks, we benefited from a geometric calculation method to track the social distance between people. Experimental results showed that our proposed models achieved significantly high performance with the help of our proposed datasets, since they contain a large amount of variation and they represent various cases in the real-world scenario. The cross-dataset experiments indicate the generalization capacity of our proposed models on unseen data. The proposed system can be effectively utilized to track all preventions against the transmission of COVID-19.

\section*{Acknowledgements}
The project on which this report is based was funded by the Federal Ministry of Education and Research (BMBF) of Germany under the number 01IS18040A. The authors are responsible for the content of this publication.

{\small
\bibliographystyle{ieee_fullname}
\bibliography{egbib}

\begin{thebibliography}{10}\itemsep=-1pt

\bibitem{WHO_website}
Coronavirus disease advice for the public.
\newblock
  \url{https://www.who.int/emergencies/diseases/novel-coronavirus-2019/advice-for-public}.
\newblock Accessed: 2021-07-01.

\bibitem{WHO_website_distance}
Covid-19: physical distancing.
\newblock
  \url{https://www.who.int/westernpacific/emergencies/covid-19/information/physical-distancing}.
\newblock Accessed: 2021-07-01.

\bibitem{kaggle853}
Face mask detection.
\newblock \url{https://www.kaggle.com/andrewmvd/face-mask-detection}.
\newblock Accessed: 2021-07-01.

\bibitem{ahmed2021deep}
Imran Ahmed, Misbah Ahmad, Joel~JPC Rodrigues, Gwanggil Jeon, and Sadia Din.
\newblock A deep learning-based social distance monitoring framework for
  covid-19.
\newblock {\em Sustainable Cities and Society}, 2021.

\bibitem{anwar2020masked}
Aqeel Anwar and Arijit Raychowdhury.
\newblock Masked face recognition for secure authentication.
\newblock {\em arXiv preprint arXiv:2008.11104}, 2020.

\bibitem{batagelj2021correctly}
Borut Batagelj, Peter Peer, Vitomir {\v{S}}truc, and Simon Dobri{\v{s}}ek.
\newblock How to correctly detect face-masks for covid-19 from visual
  information?
\newblock {\em Applied Sciences}, 11(5):2070, 2021.

\bibitem{beyan2020analysis}
Cigdem Beyan, Matteo Bustreo, Muhammad Shahid, Gian~Luca Bailo, Nicolo
  Carissimi, and Alessio Del~Bue.
\newblock Analysis of face-touching behavior in large scale social interaction
  dataset.
\newblock In {\em ICMI}, 2020.

\bibitem{cabani2021maskedface}
Adnane Cabani et~al.
\newblock Maskedface-net--a dataset of correctly/incorrectly masked face images
  in the context of covid-19.
\newblock {\em Smart Health}, 2021.

\bibitem{chen2020deep}
Jun Chen et~al.
\newblock Deep learning-based model for detecting 2019 novel coronavirus
  pneumonia on high-resolution computed tomography.
\newblock {\em Scientific reports}, 2020.

\bibitem{chen2020efficient}
Senqiu Chen, Wenbo Liu, and Gong Zhang.
\newblock Efficient transfer learning combined skip-connected structure for
  masked face poses classification.
\newblock {\em IEEE Access}, 8, 2020.

\bibitem{chowdary2020face}
G~Jignesh Chowdary, Narinder~Singh Punn, Sanjay~Kumar Sonbhadra, and Sonali
  Agarwal.
\newblock Face mask detection using transfer learning of inceptionv3.
\newblock In {\em Intl. Conf. on Big Data Analytics}, 2020.

\bibitem{damer2020effect}
Naser Damer et~al.
\newblock The effect of wearing a mask on face recognition performance: an
  exploratory study.
\newblock In {\em BIOSIG}, 2020.

\bibitem{deng2009imagenet}
Jia Deng, Wei Dong, Richard Socher, Li-Jia Li, Kai Li, and Li Fei-Fei.
\newblock Imagenet: A large-scale hierarchical image database.
\newblock In {\em CVPR}, pages 248--255. IEEE, 2009.

\bibitem{Deng_2020_CVPR}
Jiankang Deng, Jia Guo, Evangelos Ververas, Irene Kotsia, and Stefanos
  Zafeiriou.
\newblock Retinaface: Single-shot multi-level face localisation in the wild.
\newblock In {\em CVPR}, pages 5203--5212, 2020.

\bibitem{farooq2020covid}
Muhammad Farooq and Abdul Hafeez.
\newblock Covid-resnet: A deep learning framework for screening of covid19 from
  radiographs.
\newblock {\em arXiv preprint arXiv:2003.14395}, 2020.

\bibitem{golwalkar2020age}
Rucha Golwalkar and Ninad Mehendale.
\newblock Age detection with face mask using deep learning and facemasknet-9.
\newblock {\em Available at SSRN 3733784}, 2020.

\bibitem{resnetmodel}
Kaiming He, Xiangyu Zhang, Shaoqing Ren, and Jian Sun.
\newblock Deep residual learning for image recognition.
\newblock In {\em CVPR}, pages 770--778, 2016.

\bibitem{LFW}
Gary~B Huang and Erik Learned-Miller.
\newblock Labeled faces in the wild: Updates and new reporting procedures.
\newblock {\em Dept. Comput. Sci., Univ. Massachusetts Amherst, Amherst, MA,
  USA, Tech. Rep}, 14(003), 2014.

\bibitem{jiang2020retinamask}
Mingjie Jiang and Xinqi Fan.
\newblock Retinamask: a face mask detector.
\newblock {\em arXiv preprint arXiv:2005.03950}, 2020.

\bibitem{joshi2020deep}
Aniruddha~Srinivas Joshi, Shreyas~Srinivas Joshi, Goutham Kanahasabai, Rudraksh
  Kapil, and Savyasachi Gupta.
\newblock Deep learning framework to detect face masks from video footage.
\newblock In {\em CICN}, pages 435--440. IEEE, 2020.

\bibitem{karras2019style}
Tero Karras, Samuli Laine, and Timo Aila.
\newblock A style-based generator architecture for generative adversarial
  networks.
\newblock In {\em CVPR}, pages 4401--4410, 2019.

\bibitem{le1989handwritten}
Yann Le~Cun et~al.
\newblock Handwritten digit recognition with a back-propagation network.
\newblock In {\em NeurIPS}, 1989.

\bibitem{li2020using}
Lin Li et~al.
\newblock Using artificial intelligence to detect covid-19 and
  community-acquired pneumonia based on pulmonary ct: evaluation of the
  diagnostic accuracy.
\newblock {\em Radiology}, 296(2), 2020.

\bibitem{liu2016ssd}
Wei Liu, Dragomir Anguelov, Dumitru Erhan, Christian Szegedy, Scott Reed,
  Cheng-Yang Fu, and Alexander~C Berg.
\newblock Ssd: Single shot multibox detector.
\newblock In {\em ECCV}, pages 21--37. Springer, 2016.

\bibitem{celebahq}
Ziwei Liu, Ping Luo, Xiaogang Wang, and Xiaoou Tang.
\newblock Deep learning face attributes in the wild.
\newblock In {\em ICCV}, pages 3730--3738, 2015.

\bibitem{loey2021fighting}
Mohamed Loey, Gunasekaran Manogaran, Mohamed Hamed~N Taha, and Nour Eldeen~M
  Khalifa.
\newblock Fighting against covid-19: A novel deep learning model based on
  yolo-v2 with resnet-50 for medical face mask detection.
\newblock {\em Sustainable Cities and Society}, 2021.

\bibitem{loey2021hybrid}
Mohamed Loey, Gunasekaran Manogaran, Mohamed Hamed~N Taha, and Nour Eldeen~M
  Khalifa.
\newblock A hybrid deep transfer learning model with machine learning methods
  for face mask detection in the era of the covid-19 pandemic.
\newblock {\em Measurement}, 167, 2021.

\bibitem{nagrath2021ssdmnv2}
Preeti Nagrath, Rachna Jain, Agam Madan, Rohan Arora, Piyush Kataria, and Jude
  Hemanth.
\newblock Ssdmnv2: A real time dnn-based face mask detection system using
  single shot multibox detector and mobilenetv2.
\newblock {\em Sustainable cities and society}, 2021.

\bibitem{narin2020automatic}
Ali Narin, Ceren Kaya, and Ziynet Pamuk.
\newblock Automatic detection of coronavirus disease (covid-19) using x-ray
  images and deep convolutional neural networks.
\newblock {\em arXiv preprint arXiv:2003.10849}, 2020.

\bibitem{petrovic2020iot}
Nenad Petrovi{\'c} and {\DJ}or{\dj}e Koci{\'c}.
\newblock Iot-based system for covid-19 indoor safety monitoring.
\newblock {\em preprint), IcETRAN}, 2020.

\bibitem{rezaei2020deepsocial}
Mahdi Rezaei and Mohsen Azarmi.
\newblock Deepsocial: Social distancing monitoring and infection risk
  assessment in covid-19 pandemic.
\newblock {\em Applied Sciences}, 10(21):7514, 2020.

\bibitem{mobilenetv2model}
Mark Sandler, Andrew Howard, Menglong Zhu, Andrey Zhmoginov, and Liang-Chieh
  Chen.
\newblock Mobilenetv2: Inverted residuals and linear bottlenecks.
\newblock In {\em CVPR}, pages 4510--4520, 2018.

\bibitem{sathyamoorthy2020covid}
Adarsh~Jagan Sathyamoorthy, Utsav Patel, Yash~Ajay Savle, Moumita Paul, and
  Dinesh Manocha.
\newblock Covid-robot: Monitoring social distancing constraints in crowded
  scenarios.
\newblock {\em arXiv preprint arXiv:2008.06585}, 2020.

\bibitem{selvaraju2017grad}
Ramprasaath~R Selvaraju, Michael Cogswell, Abhishek Das, Ramakrishna Vedantam,
  Devi Parikh, and Dhruv Batra.
\newblock Grad-cam: Visual explanations from deep networks via gradient-based
  localization.
\newblock In {\em ICCV}, pages 618--626, 2017.

\bibitem{inceptionv3model}
Christian Szegedy, Vincent Vanhoucke, Sergey Ioffe, Jon Shlens, and Zbigniew
  Wojna.
\newblock Rethinking the inception architecture for computer vision.
\newblock In {\em CVPR}, pages 2818--2826, 2016.

\bibitem{efficientnetmodel}
Mingxing Tan and Quoc Le.
\newblock Efficientnet: Rethinking model scaling for convolutional neural
  networks.
\newblock In {\em ICML}, 2019.

\bibitem{waibel1989phoneme}
Alex Waibel, Toshiyuki Hanazawa, Geoffrey Hinton, Kiyohiro Shikano, and Kevin~J
  Lang.
\newblock Phoneme recognition using time-delay neural networks.
\newblock {\em IEEE Trans. on Acoustics, Speech, and Signal Processing},
  37(3):328--339, 1989.

\bibitem{deepHRNet}
Jingdong Wang et~al.
\newblock Deep high-resolution representation learning for visual recognition.
\newblock {\em IEEE Trans. on PAMI}, 2020.

\bibitem{wang2020masked}
Zhongyuan Wang et~al.
\newblock Masked face recognition dataset and application.
\newblock {\em arXiv preprint arXiv:2003.09093}, 2020.

\bibitem{wang2021wearmask}
Zekun Wang, Pengwei Wang, Peter~C Louis, Lee~E Wheless, and Yuankai Huo.
\newblock Wearmask: Fast in-browser face mask detection with serverless edge
  computing for covid-19.
\newblock {\em arXiv preprint arXiv:2101.00784}, 2021.

\bibitem{yang2020vision}
Dongfang Yang, Ekim Yurtsever, Vishnu Renganathan, Keith~A Redmill, and
  {\"U}mit {\"O}zg{\"u}ner.
\newblock A vision-based social distancing and critical density detection
  system for covid-19.
\newblock {\em arXiv preprint arXiv:2007.03578}, pages 24--25, 2020.

\bibitem{yang2016wider}
Shuo Yang, Ping Luo, Chen-Change Loy, and Xiaoou Tang.
\newblock Wider face: A face detection benchmark.
\newblock In {\em CVPR}, pages 5525--5533, 2016.

\bibitem{mtcnn_paper}
Kaipeng Zhang, Zhanpeng Zhang, Zhifeng Li, and Yu Qiao.
\newblock Joint face detection and alignment using multitask cascaded
  convolutional networks.
\newblock {\em IEEE Signal Proc. Letters}, 23(10), 2016.

\end{thebibliography}
}

\end{document}